# Efficient Stereo Depth Estimation for Pseudo LiDAR: A Self-Supervised Approach Based on Multi-Input ResNet Encoder

Sabir Hossain
Faculty of Engineering and Applied Science,
Ontario Tech University, Canada
sabir.hossain@ontariotechu.net

Xianke Lin*
Faculty of Engineering and Applied Science,
Ontario Tech University, Canada
xianke.lin@ontariotechu.ca

*Abstract*— Perception and localization are essential for autonomous delivery vehicles, mostly estimated from 3D LiDAR sensors due to their precise distance measurement capability. This paper presents a strategy to obtain the real-time pseudo point cloud instead of the laser sensor from the image sensor. We propose an approach to use different depth estimators to obtain pseudo point clouds like LiDAR to obtain better performance. Moreover, the training and validating strategy of the depth estimator has adopted stereo imagery data to estimate more accurate depth estimation as well as point cloud results. Our approach to generating depth maps outperforms on KITTI benchmark while yielding point clouds significantly faster than other approaches.

## I. Introduction

Understanding the three-dimension structure of the environment is possible for humans due to biological vision. Depth perception using computer vision technology is still one of the unsolved problems and most challenging issues in this research area. More significantly, proper depth perception is required for an autonomous system like an autonomous delivery vehicle. It is possible to obtain such perception from the lidar point cloud; however, lidar is a very costly technology. It will drastically increase the production cost of a delivery robot system [1]. Without a doubt, a depth-predicting system is required to find the obstacle location and avoid the collision. Many researchers have already discussed the idea of alternative LiDAR solutions due to the cost and over-dependency leading to safety risks. For example, the PSMNet [2] model defined in the pseudo-lidar paper is an image-based approach. The model architecture is too heavy that it requires more time to produce depth estimation. Therefore, the corresponding point cloud generation will be slower than LiDAR hardware, which is 10hz. Our approach uses self-supervised stereo monodepth2 [3], which is improvised to perform network training with stereo pairs in Kitti benchmark [4] datasets. Then, we used the generated disparity information to create the point cloud messages (shown in Figure 1). The main contribution of this paper –

- Adopting U-net [5] based encoder-decoder architecture as a depth network instead of the heavy PSMNet model to increase the real-time performance

- Modification in encoder network for training step. The result outperforms all the modes used by Monodepth2.

To evaluate our claim, we used a similar evaluation benchmark and produced the result that shows the superior performance of depth estimation. Then, we used the model to generate the point cloud and calculated the processing time it took to generate the depth map. Clearly, the results show that the approaches we took provide sufficient FPS to execute the whole operation in real-time.

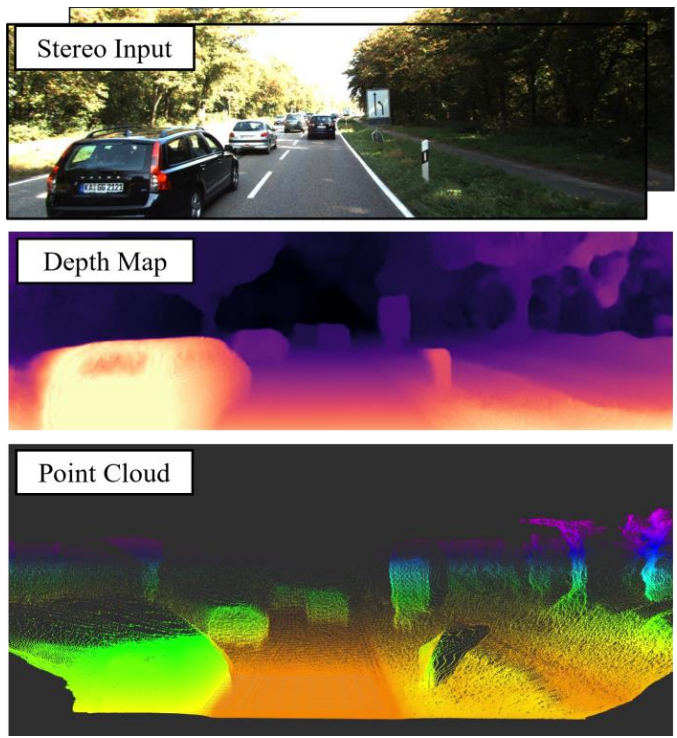

Figure 1: Point cloud generation from the depth map. Top image: Input to the pipeline- rectified stereo images from KITTI; Middle image: Estimated depth map using U-net based Depth network; Bottom image: Pseudo point-cloud generation from the disparity

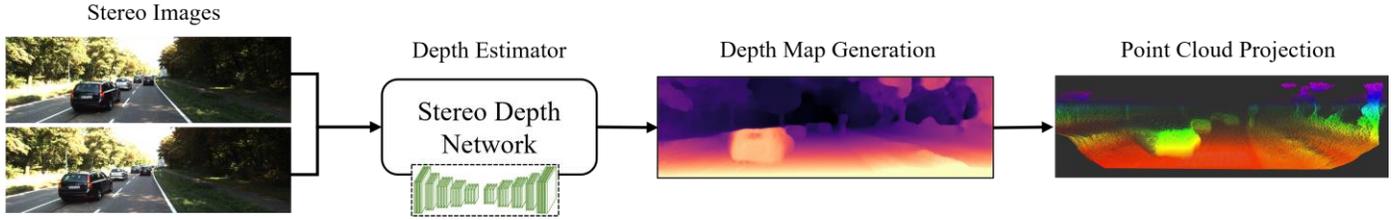

Figure 2: The proposed pipeline to generate pseudo point cloud-like LiDAR. From the stereo images, depth map prediction is made using a modified stereo depth network, then back projecting the pixel to a 3D point coordinate system

## II. RELATED WORK

The image-based depth estimation to perform perception or localization tasks can be achieved using monocular vision [6] or stereo vision [2]. An algorithm like DORN [7] outperforms by estimating the depth with low errors from the previous works on monocular depth estimation [8] [9] [10]. On the other hand, stereo-based depth prediction systems [2] show more precision in estimating disparity. However, a promising solution requires a real-time operating speed with more efficiency.

Recent studies are leveraging deep neural networks to learn model priors using pictorial depth such as texture density or object perspective directly from the training data [11]. Several technical breakthroughs in the past few years have made it possible to improve depth estimation due to the ground truth depth dataset. If the ground truth depth is not available, a possible alternative is to train models using image reconstruction. Here the model is fed either monocular temporal frames or stereo pairs of images as input. The model is trained by reducing the error in image reconstruction by imitating the depth and projecting it into a nearby view. Stereo pair is one form of self-supervision. Since stereo pair of data is available and easy to obtain, a deep network can be trained to perform depth estimation using synchronized stereo pairs during training. For the problem of novel view synthesis, the authors proposed a model with discretized depth [12] and a model predicting continuous disparity [6]. Several advancements have also been made in stereo-based approaches, including generative adversarial networks [13] and supervised datasets [13]. Also, there are approaches to predicting depth with minimized photometric reprojection error, with the use of relative pose from the source image with respect to the target image [3]. In their stereo approach, the authors used stereo pairs to calculate the losses; however, the neural architecture does not obtain the feature of other image pairs.

We demonstrate that the existing depth estimation model can be adapted to generate higher quality results by combining the stereo pair in input layers rather than using the pair to calculate relative pose loss only. Moreover, we used the modified model to generate point clouds in real-time.

## III. METHOD

This section introduces the architecture of our modified deep network; then presents the strategy to split, point cloud generation, post-processing steps and evaluating techniques used to compare. The proposed pipeline is shown in figure 2, and the modules are discussed in detail in this section.

### A. Stereo Training using Depth Network

The architecture is encoder-decoder-based classic U-Net (shown in Figure 3). The encoder is a pre-train ResNet model [14], and the decoder converts the sigmoid output to a depth map. U-Net architecture merges various scale features with varying receptive field sizes and concatenates the feature maps after upsampling them by pooling them into distinct sizes. The ResNet encoder module usually accepts single RGB images as input. The ResNet encoder is modified to accept a pair of stereo frames, or six channels, as input for the posture model. As a result, instead of the ResNet default of (3, 192, 640), the ResNet encoder uses convolutional weights in the first layer of shape (6, 192, 640). The depth decoder is a fully convolutional network that takes advantage of feature maps of different scales and concatenates them after upsampling. There is a sigmoid activation at the last layer that outputs a normalized disparity map between 0 to 1. Table I shows the output total number of trainable parameters for encoder are 11,185,920 for (192,640) size of image input, whereas, for single image layer based encoder, it would be 11,176,512. The ResNet encoder has 20

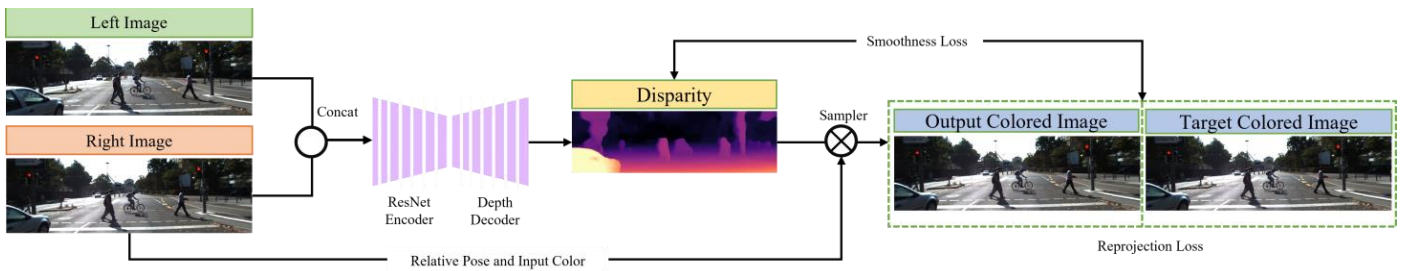

Figure 3: The modified approach with stereo pair in the encoder architecture. Loss is calculated based on the other pair

TABLE I
MODEL SUMMERY FOR ENCODER MODULE

| Layer (type:depth-idx) | Output Shape | Param |
|---|---|---|
| Conv2D:1-1 | [1, 64, 96, 320] | 18,816 |
| BatchNorm2d: 1-2 | [1, 64, 96, 320] | 128 |
| ReLU: 1-3 | [1, 64, 96, 320] | -- |
| MaxPool2d: 1-4 | [1, 64, 48, 160] | -- |
| Sequential: 1-5 | [1, 64, 48, 160] | -- |
|     BasicBlock: 2-1 | [1, 64, 48, 160] | 73,984 |
|     BasicBlock: 2-2 | [1, 64, 48, 160] | 73,984 |
| Sequential: 1-6 | [1, 128, 24, 80] | -- |
|     BasicBlock: 2-3 | [1, 128, 24, 80] | 230,144 |
|     BasicBlock: 2-4 | [1, 128, 24, 80] | 295,424 |
| Sequential: 1-7 | [1, 256, 12, 40] | -- |
|     BasicBlock: 2-5 | [1, 256, 12, 40] | 919,040 |
|     BasicBlock: 2-6 | [1, 256, 12, 40] | 1,180,672 |
| Sequential: 1-8 | [1, 512, 6, 20] | -- |
|     BasicBlock: 2-7 | [1, 512, 6, 20] | 3,673,088 |
|     BasicBlock: 2-8 | [1, 512, 6, 20] | 4,720,640 |
| | Total params: 11,185,920 | |
| | Trainable params: 11,185,920 | |
| | Non-trainable params: 0 | |

Conv2d layers, 20 BatchNom2D layers, 17 ReLU, 1 MaxPool2D layer, and eight basic blocks in total. The decoder layer has the same block, kernel size and strides.

In monocular mode, monodepth2 authors used temporal frame in Posenet [3] instead of stereo pair to calculate the extrinsic parameter of the camera and the pose of the image frame. Our approach will not rely on temporal frames for self-supervised prediction. The reprojection loss is calculated using SSIM [15] between prediction and target in stereo mode in stereo training. The metric reprojection error $L_p$ is calculated from the relative pose $T_{s \to t}$ of source view denoted as $I_s$ with respect to its target image $I_t$. In our training, the other stereo pair will provide the relative position $T_{s \to t}$ of the source image $I_s$. This rotation and translation information will be used to calculate the mapping from the source frame to the target frame. Simultaneously, the ResNet architecture is fed with both image pairs (shown in Figure 3). The other can be considered the stereo pair of source images by considering one as the primary input. The target image is reprojected from the predicted depth and transformation matrix from the stereo pair using the intrinsic matrix. Then the method used bilinearly sampling to sample the source image from the target image. This loss aims to minimize the difference between the target picture and the reconstructed target image, in which depth is the most crucial factor. Instead of averaging the photometric error across all source frames, the method utilized the minimum at each pixel. The equation of photometric loss $L_p$ can be represented [3] like the following equation (1)

$$L_p = \min_s RE(I_t, I_{s \to t}) \quad (1)$$

Here $RE$ is the metric reconstruction error. $I_{t' \to t}$ is obtained [16] from the projected depth $D$, intrinsic parameter $K$ and relative pose like the following equation (2). $\langle \rangle$ is the bilinear sampling operator and $proj()$ denotes 2D-cordinate of projected image.

$$I_{s \to t} = I_s \langle proj(K, D, T_{s \to t}) \rangle \quad (2)$$

On the other hand, the edge-aware smoothness loss $L_s$ is also calculated between the target frame and mean-normalized inversed depth value. It boosts the model to recognize the sharp edges and eliminate noises. The following equation (3) is the final training loss function

$$L = \mu L_p + \lambda L_s \quad (3)$$

Where μ is the mask pixel value which is $\mu \in \{0,1\}$ obtained from the auto-masking method [3], and $\lambda$ is the smoothness term that is 0.001. Learning rate $10^{-4}$, batch size 12, epochs size 20 is used while training model size of both $640 \times 192$ and $1024 \times 320$.

B. Dataset Splitting

We use the data split of Zhou et al.'s [16], which has 39,810 datasets for training and 4,424 for validation. The intrinsic parameter provided by KITTI, which includes focal length and image center, is normalized with respect to image resolution. A horizontal translation of fixed size is applied to the horizontal transformation between stereo frames. The neural network is fed the image from the split file along with the corresponding pair. However, for rest of the calculation is based on the first taken from the split dataset, not the pair image. In stereo supervision,

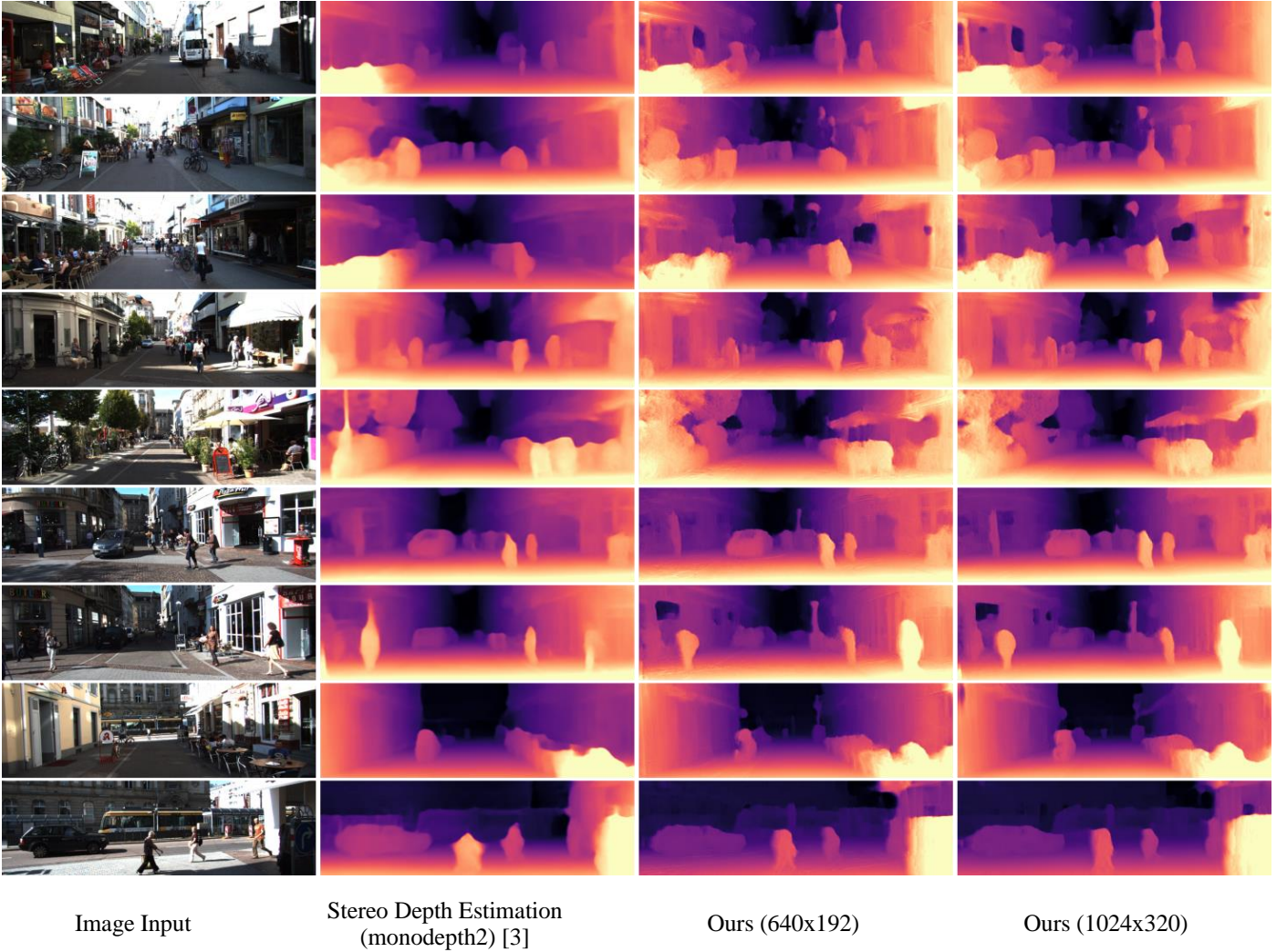

Figure 4: Qualitative results on the KITTI scene. The first column denotes the primary image input; the second column has produced the result from the stereo depth estimator by monodepth2; the third and fourth column is produced from our results 640x192 and 1024x320 model, respectively.

median scaling is not needed as the camera baseline can be used as a reference for scale.

*C. Pointcloud Back-Projection*

The depth (*z*) can be obtained from a stereo disparity estimation system that requires pair of right-left images with a horizontal baseline *b*. The depth estimation system will consider the left image as a reference and save the disparity map *d* with respect to the right image for each pixel $(x, y)$. Considering the focal length of the camera, *f*, the following equation (4) of depth transformation can be obtained,

$$z(x,y) = \frac{b \times f}{d(x,y)} \qquad (4)$$

Points clouds have their own 3D coordinate with respect to a reference viewpoint and direction. Such 3D coordinate can be obtained by back-projecting all the depth pixels to a 3-dimensional coordinate system that will contain the points coordinate as $[(X_n, Y_n, Z_n)]_{n=0}^{N}$, $N$ is the number of total points generated from the depth pixel. The back-projection was performed on the KITTI dataset images using their project matrices. The 3D location of each point can be obtained using the following equation (5-7) with respect to the left cameras frame reference that can be calculated from the calibration matrices.

$$width, X(x,y) = \frac{z \times (x - c_x)}{f_x} \qquad (5)$$

$$height, Y(x,y) = \frac{z \times (y - c_y)}{f_y} \qquad (6)$$

$$depth, Z(x,y) = z \quad (7)$$

Where $f$ is the focal length of the camera and $(c_x, c_y)$ is the center pixel location of the image. Similar steps of back-projection are used to generate a pseudo-Lidar point cloud [17].

### D. Post-Processing Step

The method can adopt a post-processing step while training to achieve a significant accurate result in the evolution benchmark step. In order to obtain the model with the post-processing step, the stereo network is trained with the images two times, flipped and un-flipped. An unsupervised depth estimator introduces this type of two-forward pass-through network technique to improve the result [6].

### E. Evaluation Metric

Abs Rel, Sq Rel, RMSE, and RMSE log means absolute error, squared error, linear root mean squared error, and logarithmic root mean squared error, respectively, between ground truth and prediction. These values indicate the lower, the better result. On the other hand, δ<x denotes the ratio prediction and ground truth between x and $1/x$. The results which are closer to 1 are better results. Instead of LiDAR reproject, ground truth depth from the KITTI depth prediction dataset [37] is used to evaluate the prediction method. During the evaluation of our method, we used the same ground truth mentioned by monodepth2 [3] while using stereo images as input in the encoder's input layer.

TABLE II.
QUANTITATIVE RESULTS WITH ALL VARIANTS OF MONODEPTH2 AND OTHER SELF-SUPERVISED METHODS ON THE KITTI 2015 DATASET

| Method | Train Input | Abs Rel | Sq Rel | RMSE | RMSE log | $\delta < 1.25$ | $\delta < 1.25^2$ | $\delta < 1.25^3$ |
|---|---|---|---|---|---|---|---|---|
| Eigen 2014 [9] | D | 0.203 | 1.548 | 6.307 | 0.282 | 0.702 | 0.89 | 0.89 |
| Liu 2005 [8] | D | 0.201 | 1.584 | 6.471 | 0.273 | 0.68 | 0.898 | 0.967 |
| Klodt 2018 [18] | D*M | 0.166 | 1.49 | 5.998 | - | 0.778 | 0.919 | 0.966 |
| AdaDepth 2018 [10] | D* | 0.167 | 1.257 | 5.578 | 0.237 | 0.771 | 0.922 | 0.971 |
| Kuznietsov 2017 [19] | DS | 0.113 | 0.741 | 4.621 | 0.189 | 0.862 | 0.96 | 0.986 |
| DVSO 2018 [20] | D*S | 0.097 | 0.734 | 4.442 | 0.187 | 0.888 | 0.958 | 0.98 |
| SVSM FT 2018 [13] | DS | 0.094 | 0.626 | 4.252 | 0.177 | 0.891 | 0.965 | 0.984 |
| Guo 2018 [21] | DS | 0.096 | 0.641 | 4.095 | 0.168 | 0.892 | 0.967 | 0.986 |
| DORN 2018 [22] | D | 0.072 | 0.307 | 2.727 | 0.12 | 0.932 | 0.984 | 0.994 |
| Zhou 2017 [16] | M | 0.183 | 1.595 | 6.709 | 0.27 | 0.734 | 0.902 | 0.959 |
| Yang 2018 [7] | M | 0.182 | 1.481 | 6.501 | 0.267 | 0.725 | 0.906 | 0.963 |
| Mahjourian 2018 [23] | M | 0.163 | 1.24 | 6.22 | 0.25 | 0.762 | 0.916 | 0.968 |
| GeoNet 2018 [24] | M | 0.149 | 1.06 | 5.567 | 0.226 | 0.796 | 0.935 | 0.975 |
| DDVO 2018 [25] | M | 0.151 | 1.257 | 5.583 | 0.228 | 0.81 | 0.936 | 0.974 |
| DF-Net 2018 [26] | M | 0.15 | 1.124 | 5.507 | 0.223 | 0.806 | 0.933 | 0.973 |
| LEGO 2018 [27] | M | 0.162 | 1.352 | 6.276 | 0.252 | - | - | - |
| Ranjan 2019 [28] | M | 0.148 | 1.149 | 5.464 | 0.226 | 0.815 | 0.935 | 0.973 |
| EPC++ 2020 [29] | M | 0.141 | 1.029 | 5.35 | 0.216 | 0.816 | 0.941 | 0.976 |
| Struct2depth '(M)' 2019 [30] | M | 0.141 | 1.026 | 5.291 | 0.215 | 0.816 | 0.945 | 0.979 |
| Monodepth2 w/o pretraining 2019 [3] | M | 0.132 | 1.044 | 5.142 | 0.21 | 0.845 | 0.948 | 0.977 |
| Monodepth2 (640x192), 2019 [3] | M | 0.115 | 0.903 | 4.863 | 0.193 | 0.877 | 0.959 | 0.981 |
| Monodepth2 (1024 x 320), 2019 [3] | M | 0.115 | 0.882 | 4.701 | 0.19 | 0.879 | 0.961 | 0.982 |
| Garg 2016 [31] | S | 0.152 | 1.226 | 5.849 | 0.246 | 0.784 | 0.921 | 0.967 |
| Monodepth R50 2017 [6] | S | 0.133 | 1.142 | 5.533 | 0.23 | 0.83 | 0.936 | 0.97 |
| StrAT 2018 [32] | S | 0.128 | 1.019 | 5.403 | 0.227 | 0.827 | 0.935 | 0.971 |
| 3Net (R50), 2018 [33] | S | 0.129 | 0.996 | 5.281 | 0.223 | 0.831 | 0.939 | 0.974 |
| 3Net (VGG), 2018 [33] | S | 0.119 | 1.201 | 5.888 | 0.208 | 0.844 | 0.941 | 0.978 |
| SuperDepth+PP, 2019 [34] (1024x382) | S | 0.112 | 0.875 | 4.958 | 0.207 | 0.852 | 0.947 | 0.977 |
| Monodepth2 w/o pretraining 2019 [3] | S | 0.13 | 1.144 | 5.485 | 0.232 | 0.831 | 0.932 | 0.968 |
| Monodepth2 (640x192), 2019 [3] | S | 0.109 | 0.873 | 4.96 | 0.209 | 0.864 | 0.948 | 0.975 |
| Monodepth2 (1024x320) 2019 [3] | S | 0.107 | 0.849 | 4.764 | 0.201 | 0.874 | 0.953 | 0.977 |
| UnDeepVO, 2018 [35] | MS | 0.183 | 1.73 | 6.57 | 0.268 | - | - | - |
| Zhan FullNYU, 2018 [36] | D*MS | 0.135 | 1.132 | 5.585 | 0.229 | 0.82 | 0.933 | 0.971 |
| EPC++, 2020 [29] | MS | 0.128 | 0.935 | 5.011 | 0.209 | 0.831 | 0.945 | 0.979 |
| Monodepth2 w/o pretraining, 2019 [3] | MS | 0.127 | 1.031 | 5.266 | 0.221 | 0.836 | 0.943 | 0.974 |
| Monodepth2 (640x192), 2019 [3] | MS | 0.106 | 0.818 | 4.75 | 0.196 | 0.874 | 0.957 | 0.979 |
| Monodepth2 (1024x320), 2019 [3] | MS | 0.106 | 0.806 | 4.63 | 0.193 | 0.876 | 0.958 | 0.98 |
| **Ours w/o pretraining (640x192)** | **S** | **0.083** | **0.768** | **4.467** | **0.185** | **0.911** | **0.959** | **0.977** |
| **Ours (640x192)** | **S** | **0.080** | **0.747** | **4.346** | **0.181** | **0.918** | **0.961** | **0.978** |
| **Ours (1024x320)** | **S** | **0.077** | **0.723** | **4.233** | **0.179** | **0.922** | **0.961** | **0.978** |
| **Ours (1024x320) + PP** | **S** | **0.075** | **0.700** | **4.196** | **0.176** | **0.924** | **0.963** | **0.979** |

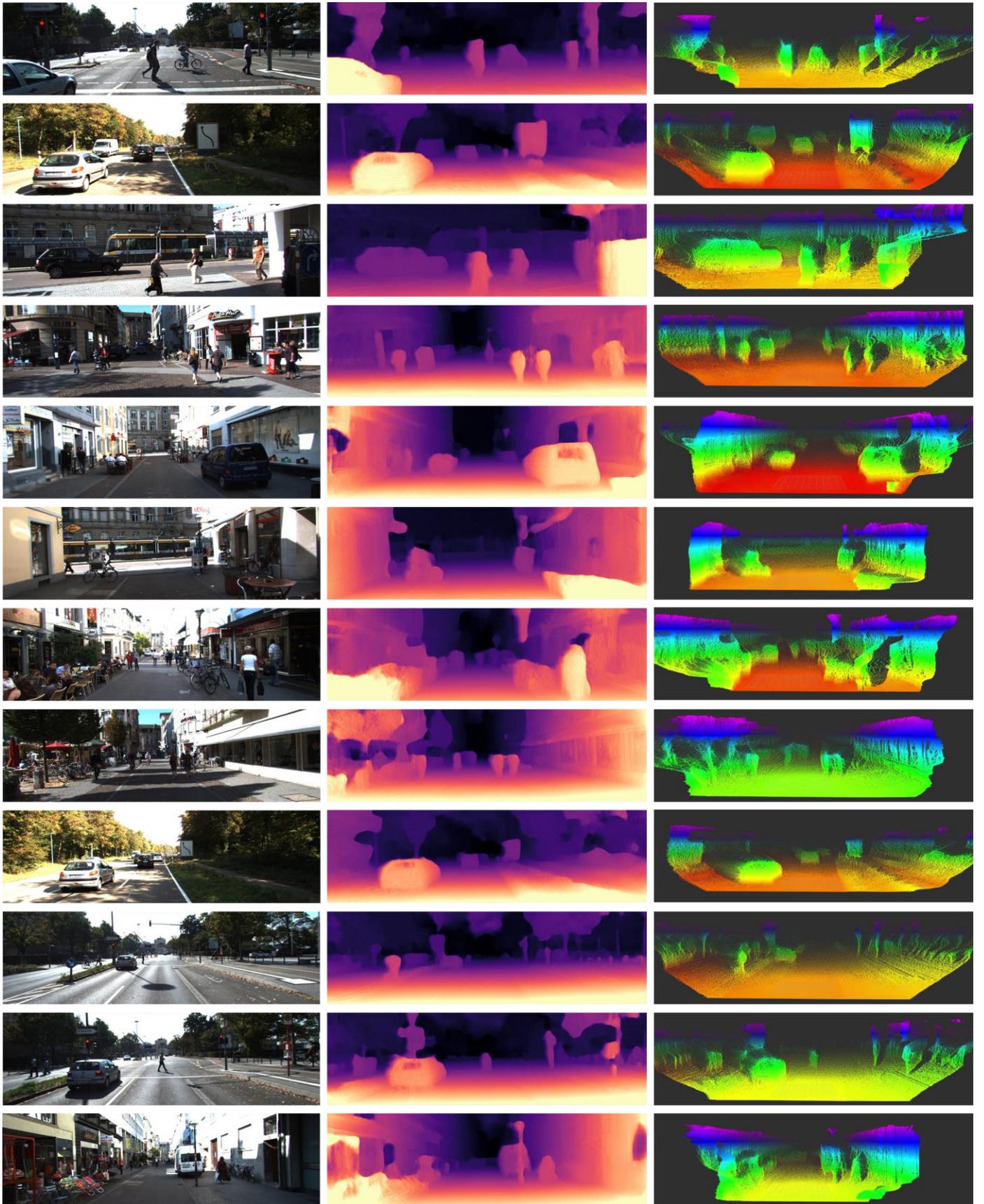

Figure 5: Visualization of Point Cloud generated from the depth estimation; the first column is the input pairs to the neural network; the second column is depth prediction, and the third column presents the point cloud result.

## IV. EXPERIMENT & RESULTS

Figure 4 presents the qualitative results on a specific KITTI scene. The first, second, and 7th-row results show that our method adequately recognizes the pole. Also, other results show the size of pedestrians (for example, result in row 9), the shape of objects (for example, result in rows 6 and 8) and building (for example, result in rows 5 and 7) are more aligned with the original image. From this visual result, it is clear that our depth estimator can predict some of the features like poles or street signs, moving objects and objects at far distances. The comparison is performed with other Monodepth2 modes: monocular only (M), stereo-only (S), and both (MS), along with other self-supervised models presented in the paper [3]. Table II shows that our method (highlighted with bold font) outperforms all the variants, including self-supervised methods, except DORN [21]. Here D refers to depth supervision, D* refers to auxiliary depth supervision, M refers to self-supervised mono supervision, S refers to self-supervised stereo supervision, and PP refers to post-processing. The result achieving higher accuracy is due to the introduction of stereo pairs in the input layer of ResNet architecture. We used a common system to compare average processing speed (11th Gen Intel i7-11800H, 2.30GHz, NVIDIA GeForce RTX 3070 Laptop GPU-8GB, Ubuntu 18.04, Torch 1.10.0, CUDA 11.3). The results in Table III with ** shows more FPS than other image resolution. If the model resolution and image solution size are similar, the process does not use functional interpolation to resize the depth metrics to the image resolution. Therefore requires less time to predict the final output. Other resolutions present low FPS due to computationally expensive rescaling of the depth map. The processing time for PSMNet requires much longer than U-Net-based architecture.

The main module responsible for FPS is the depth prediction network. We presented the result with models 640x192 and 1024x320 in Table III. Using stereo model 1024x320, we can obtain higher accuracy in real-time. Figure 5 shows the point cloud visualization result on KITTI scenes. We used ROS converted bag of KITTI dataset and RViz to visualize the point cloud data. The point cloud visualization shows the perception of pedestrians, bicycles, cars, and traffic signs in 3D space.

## V. CONCLUSION

In this paper, we have presented a strategy aimed at reducing the gap between point clouds from real LiDAR devices and image-based point clouds. We presented performance results for operating the model with U-net architecture. Both versions of our resolution (avg 57.2 and 31.9 FPS, respectively) indicate real-time operating performance. Moreover, we improved the network input layer by introducing stereo pairs to the input layer. Improvement in the stereo-based network is due to stereo information that helps the network to conceive more perceptions about moving and standstill objects. The final result shows more accurate results on the model with 1024x320 and post-processing-based training on 1024x320. The improvement we achieved from the modified network is comparatively greater than its previous versions. Initially, we took a different approach to introduce more pixel features to the model. We tried to concatenate the temporal frames in the input layers; however, the result was poor. Later, we adopted the approach of stereo pairs since the model has no experience with stereo pairs. More significantly, the LiDAR device is the most expensive commercial component for delivery vehicles. The image-based approach is the way to close this gap in cost. In future work, we aim to perform 3D object detection and SLAM algorithm over the point cloud achieved from depth prediction.

TABLE III
AVERAGE PROCESSING SPEED

| Method | Image Resolution | FPS (Avg) |
|---|---|---|
| PSMNet [2] [17] | 720 x 480 | 2.857 |
| | 1080 x 720 | 1.298 |
| | 640x256 | 6.25 |
| | 1024 x 320 | 3.125 |
| Stereo Depth Estimation, Monodepth2 (640x192) [3] | 720 x 480 | 40.6 |
| | 1080 x 720 | 21.7 |
| | 640x192 | 60.9** |
| | 1024 x 320 | 42.9 |
| Ours (640x192) | 720 x 480 | 32.1 |
| | 1080 x 720 | 19.1 |
| | **640x192** | **57.2**** |
| | 1024 x 320 | 34.2 |
| Ours (1024x320) | 720 x 480 | 23.1 |
| | 1080 x 720 | 15.64 |
| | 640x192 | 26.6 |
| | **1024 x 320** | **31.9**** |

## SUPPLEMENTARY MATERIAL

The code is available upon request to the authors. For the paper's code, contact the following email: xianke.lin@ontariotechu.ca.

## ACKNOWLEDGMENT

I would like to show my gratitude to my supervisor, Prof Xianke Lin, for providing detailed guidance and support for this project. He provided the idea of increasing features in the input layer. Moreover, throughout the year, he has shown me all the essential qualities of being a dedicated and successful researcher. Thanks to the authors of Monodepth2 authors who shared their code and results.

## REFERENCES

[1] D. Jennings and M. Figliozzi, "Study of road autonomous delivery robots and their potential effects on freight efficiency and travel," *Transp. Res. Rec.*, vol. 2674, no. 9, pp. 1019–1029, 2020, doi: 10.1177/0361198120933633.

[2] J. R. Chang and Y. S. Chen, "Pyramid Stereo Matching Network," in *Proceedings of the IEEE Computer Society Conference on Computer Vision and Pattern Recognition*, 2018, pp. 5410–5418, doi: 10.1109/CVPR.2018.00567.

[3] C. Godard, O. Mac Aodha, M. Firman, and G. Brostow, "Digging into self-supervised monocular depth estimation," in *Proceedings of the IEEE International Conference on Computer Vision*, 2019, vol. 2019-Octob, pp. 3827–3837, doi: 10.1109/ICCV.2019.00393.

[4] A. Geiger, P. Lenz, C. Stiller, and R. Urtasun, "Vision meets robotics: The KITTI dataset," *Int. J. Rob. Res.*, vol. 32, no. 11, pp. 1231–1237, 2013, doi: 10.1177/0278364913491297.

[5] O. Ronneberger, P. Fischer, and T. Brox, "U-net: Convolutional networks for biomedical image segmentation," in *Lecture Notes in Computer Science (including subseries Lecture Notes in Artificial Intelligence and Lecture Notes in Bioinformatics)*, 2015, vol. 9351, pp. 234–241, doi: 10.1007/978-3-319-24574-4_28.

[6] C. Godard, O. Mac Aodha, and G. J. Brostow, "Unsupervised monocular depth estimation with left-right consistency," in *Proceedings - 30th IEEE Conference on Computer Vision and Pattern Recognition, CVPR 2017*, 2017, vol. 2017-Janua, pp. 6602–6611, doi: 10.1109/CVPR.2017.699.

[7] Z. Yang, P. Wang, W. Xu, L. Zhao, and R. Nevatia, "Unsupervised learning of geometry from videos with edge-aware depth-normal consistency," *32nd AAAI Conf. Artif. Intell. AAAI 2018*, pp. 7493–7500, 2018.

[8] A. Saxena, S. H. Chung, and A. Y. Ng, "Learning depth from single monocular images," *Adv. Neural Inf. Process. Syst.*, vol. 18, pp. 1161–1168, 2005.

[9] D. Eigen, C. Puhrsch, and R. Fergus, "Depth map prediction from a single image using a multi-scale deep network," *Adv. Neural Inf. Process. Syst.*, vol. 27, 2014.

[10] J. N. Kundu, P. K. Uppala, A. Pahuja, and R. V. Babu, "AdaDepth: Unsupervised Content Congruent Adaptation for Depth Estimation," in *Proceedings of the IEEE Computer Society Conference on Computer Vision and Pattern Recognition*, 2018, pp. 2656–2665, doi: 10.1109/CVPR.2018.00281.

[11] S. H. M. Miangoleh, S. Dille, L. Mai, S. Paris, and Y. Aksoy, "Boosting Monocular Depth Estimation Models to High-resolution via Content-adaptive Multi-Resolution Merging," in *Proceedings of the IEEE Computer Society Conference on Computer Vision and Pattern Recognition*, 2021, pp. 9680–9689, doi: 10.1109/CVPR46437.2021.00956.

[12] J. Xie, R. Girshick, and A. Farhadi, "Deep3D: Fully automatic 2D-to-3D video conversion with deep convolutional neural networks," in *Lecture Notes in Computer Science (including subseries Lecture Notes in Artificial Intelligence and Lecture Notes in Bioinformatics)*, 2016, vol. 9908 LNCS, pp. 842–857, doi: 10.1007/978-3-319-46493-0_51.

[13] Y. Luo *et al.*, "Single View Stereo Matching," in *Proceedings of the IEEE Computer Society Conference on Computer Vision and Pattern Recognition*, 2018, pp. 155–163, doi: 10.1109/CVPR.2018.00024.

[14] K. He, X. Zhang, S. Ren, and J. Sun, "Deep residual learning for image recognition," in *Proceedings of the IEEE Computer Society Conference on Computer Vision and Pattern Recognition*, 2016, vol. 2016-December, pp. 770–778, doi: 10.1109/CVPR.2016.90.

[15] Z. Wang, A. C. Bovik, H. R. Sheikh, and E. P. Simoncelli, "Image quality assessment: From error visibility to structural similarity," *IEEE Trans. Image Process.*, vol. 13, no. 4, pp. 600–612, 2004, doi: 10.1109/TIP.2003.819861.

[16] T. Zhou, M. Brown, N. Snavely, and D. G. Lowe, "Unsupervised learning of depth and ego-motion from video," in *Proceedings - 30th IEEE Conference on Computer Vision and Pattern Recognition, CVPR 2017*, 2017, vol. 2017-Janua, pp. 6612–6621, doi: 10.1109/CVPR.2017.700.

[17] Y. Wang, W. L. Chao, Di. Garg, B. Hariharan, M. Campbell, and K. Q. Weinberger, "Pseudo-lidar from visual depth estimation: Bridging the gap in 3D object detection for autonomous driving," in *Proceedings of the IEEE Computer Society Conference on Computer Vision and Pattern Recognition*, 2019, vol. 2019-June, pp. 8437–8445, doi: 10.1109/CVPR.2019.00864.

[18] M. Klodt and A. Vedaldi, "Supervising the new with the old: Learning SFM from SFM," in *Lecture Notes in Computer Science (including subseries Lecture Notes in Artificial Intelligence and Lecture Notes in Bioinformatics)*, 2018, vol. 11214 LNCS, pp. 713–728, doi: 10.1007/978-3-030-01249-6_43.

[19] Y. Kuznietsov, J. Stückler, and B. Leibe, "Semi-supervised deep learning for monocular depth map prediction," in *Proceedings - 30th IEEE Conference on Computer Vision and Pattern Recognition, CVPR 2017*, 2017, vol. 2017-Janua, pp. 2215–2223, doi: 10.1109/CVPR.2017.238.

[20] N. Yang, R. Wang, J. Stückler, and D. Cremers, "Deep virtual stereo odometry: Leveraging deep depth prediction for monocular direct sparse odometry," in *Lecture Notes in Computer Science (including subseries Lecture Notes in Artificial Intelligence and Lecture Notes in Bioinformatics)*, 2018, vol. 11212 LNCS, pp. 835–852, doi: 10.1007/978-3-030-01237-3_50.

[21] X. Guo, H. Li, S. Yi, J. Ren, and X. Wang, "Learning Monocular Depth by Distilling Cross-Domain Stereo

Networks," in *Lecture Notes in Computer Science (including subseries Lecture Notes in Artificial Intelligence and Lecture Notes in Bioinformatics)*, 2018, vol. 11215 LNCS, pp. 506–523, doi: 10.1007/978-3-030-01252-6_30.

[22] H. Fu, M. Gong, C. Wang, K. Batmanghelich, and D. Tao, "Deep Ordinal Regression Network for Monocular Depth Estimation," in *Proceedings of the IEEE Computer Society Conference on Computer Vision and Pattern Recognition*, 2018, pp. 2002–2011, doi: 10.1109/CVPR.2018.00214.

[23] R. Mahjourian, M. Wicke, and A. Angelova, "Unsupervised Learning of Depth and Ego-Motion from Monocular Video Using 3D Geometric Constraints," in *Proceedings of the IEEE Computer Society Conference on Computer Vision and Pattern Recognition*, 2018, pp. 5667–5675, doi: 10.1109/CVPR.2018.00594.

[24] Z. Yin and J. Shi, "GeoNet: Unsupervised Learning of Dense Depth, Optical Flow and Camera Pose," in *Proceedings of the IEEE Computer Society Conference on Computer Vision and Pattern Recognition*, 2018, pp. 1983–1992, doi: 10.1109/CVPR.2018.00212.

[25] C. Wang, J. M. Buenaposada, R. Zhu, and S. Lucey, "Learning Depth from Monocular Videos Using Direct Methods," in *Proceedings of the IEEE Computer Society Conference on Computer Vision and Pattern Recognition*, 2018, pp. 2022–2030, doi: 10.1109/CVPR.2018.00216.

[26] Y. Zou, Z. Luo, and J. Bin Huang, "DF-Net: Unsupervised Joint Learning of Depth and Flow Using Cross-Task Consistency," in *Lecture Notes in Computer Science (including subseries Lecture Notes in Artificial Intelligence and Lecture Notes in Bioinformatics)*, 2018, vol. 11209 LNCS, pp. 38–55, doi: 10.1007/978-3-030-01228-1_3.

[27] Z. Yang, P. Wang, Y. Wang, W. Xu, and R. Nevatia, "LEGO: Learning Edge with Geometry all at Once by Watching Videos," in *Proceedings of the IEEE Computer Society Conference on Computer Vision and Pattern Recognition*, 2018, pp. 225–234, doi: 10.1109/CVPR.2018.00031.

[28] A. Ranjan *et al.*, "Competitive collaboration: Joint unsupervised learning of depth, camera motion, optical flow and motion segmentation," in *Proceedings of the IEEE Computer Society Conference on Computer Vision and Pattern Recognition*, 2019, vol. 2019-June, pp. 12232–12241, doi: 10.1109/CVPR.2019.01252.

[29] C. Luo *et al.*, "Every Pixel Counts ++: Joint Learning of Geometry and Motion with 3D Holistic Understanding," *IEEE Trans. Pattern Anal. Mach. Intell.*, vol. 42, no. 10, pp. 2624–2641, 2020, doi: 10.1109/TPAMI.2019.2930258.

[30] V. Casser, S. Pirk, R. Mahjourian, and A. Angelova, "Depth prediction without the sensors: Leveraging structure for unsupervised learning from monocular videos," in *33rd AAAI Conference on Artificial Intelligence, AAAI 2019, 31st Innovative Applications of Artificial Intelligence Conference, IAAI 2019 and the 9th AAAI Symposium on Educational Advances in Artificial Intelligence, EAAI 2019*, 2019, vol. 33, no. 01, pp. 8001–8008, doi: 10.1609/aaai.v33i01.33018001.

[31] R. Garg, B. G. Vijay Kumar, G. Carneiro, and I. Reid, "Unsupervised CNN for single view depth estimation: Geometry to the rescue," in *Lecture Notes in Computer Science (including subseries Lecture Notes in Artificial Intelligence and Lecture Notes in Bioinformatics)*, 2016, vol. 9912 LNCS, pp. 740–756, doi: 10.1007/978-3-319-46484-8_45.

[32] I. Mehta, P. Sakurikar, and P. J. Narayanan, "Structured adversarial training for unsupervised monocular depth estimation," in *Proceedings - 2018 International Conference on 3D Vision, 3DV 2018*, 2018, pp. 314–323, doi: 10.1109/3DV.2018.00044.

[33] M. Poggi, F. Tosi, and S. Mattoccia, "Learning monocular depth estimation with unsupervised trinocular assumptions," in *Proceedings - 2018 International Conference on 3D Vision, 3DV 2018*, 2018, pp. 324–333, doi: 10.1109/3DV.2018.00045.

[34] S. Pillai, R. Ambruș, and A. Gaidon, "Superdepth: Self-supervised, super-resolved monocular depth estimation," in *2019 International Conference on Robotics and Automation (ICRA)*, 2019, pp. 9250–9256.

[35] R. Li, S. Wang, Z. Long, and D. Gu, "UnDeepVO: Monocular Visual Odometry Through Unsupervised Deep Learning," in *Proceedings - IEEE International Conference on Robotics and Automation*, 2018, pp. 7286–7291, doi: 10.1109/ICRA.2018.8461251.

[36] H. Zhan, R. Garg, C. S. Weerasekera, K. Li, H. Agarwal, and I. M. Reid, "Unsupervised Learning of Monocular Depth Estimation and Visual Odometry with Deep Feature Reconstruction," in *Proceedings of the IEEE Computer Society Conference on Computer Vision and Pattern Recognition*, 2018, pp. 340–349, doi: 10.1109/CVPR.2018.00043.

[37] J. Uhrig, N. Schneider, L. Schneider, U. Franke, T. Brox, and A. Geiger, "Sparsity Invariant CNNs," in *Proceedings - 2017 International Conference on 3D Vision, 3DV 2017*, 2018, pp. 11–20, doi: 10.1109/3DV.2017.00012.